\documentclass[10pt,twocolumn,letterpaper]{article}

\usepackage[preprint]{cvpr}      

\usepackage{soul}
\newcommand{\gh}[1]{{\color{black}#1}}

\newcommand{\symref}[2]{%
  \hyperref[#1]{#2}%
}

\definecolor{cvprblue}{rgb}{0.21,0.49,0.74}
\usepackage[pagebackref,breaklinks,colorlinks,allcolors=cvprblue]{hyperref}
\usepackage{booktabs} 
\usepackage{amssymb} 
\usepackage{pifont}
\usepackage{multirow}
\usepackage{listings}
\usepackage{xcolor}
\usepackage{caption}

\definecolor{eclipseBlue}{RGB}{42,0.0,255}
\definecolor{eclipseGreen}{RGB}{63,127,95}
\definecolor{eclipsePurple}{RGB}{127,0,85}
\definecolor{backcolour}{rgb}{0.95,0.95,0.92}

\lstdefinelanguage{json}{
    basicstyle=\ttfamily\scriptsize,
    commentstyle=\color{eclipseGreen}, 
    keywordstyle=\color{eclipseBlue},  
    stringstyle=\color{eclipsePurple}, 
    numbers=left,
    numberstyle=\tiny\color{gray},
    stepnumber=1,
    numbersep=5pt,
    backgroundcolor=\color{backcolour},
    showstringspaces=false,
    breaklines=true,
    frame=single,
    tabsize=2,
    morestring=[b]", 
    morecomment=[l]{//}, 
    literate=
     *{:}{{{\color{black}{:}}}}{1}
      {,}{{{\color{black}{,}}}}{1}
      {\{}{{{\color{black}{\{}}}}{1}
      {\}}{{{\color{black}{\}}}}}{1}
      {[}{{{\color{black}{[}}}}{1}
      {]}{{{\color{black}{]}}}}{1},
}

\def\paperID{} 
\def\confName{CVPR}
\def\confYear{2026}
\def\OurModel{D3D-VLP}
\title{D3D-VLP: Dynamic 3D Vision-Language-Planning Model for Embodied Grounding and Navigation
}

\author{%
  Zihan Wang\textsuperscript{1} \quad Seungjun Lee\textsuperscript{1} \quad Guangzhao Dai\textsuperscript{2} \quad Gim Hee Lee\textsuperscript{1} \\
  \textsuperscript{1}School of Computing, National University of Singapore \\
  \textsuperscript{2}Nanjing University of Science and Technology \\
    {\tt zihan.wang@u.nus.edu
    }
}

\begin{document}

\maketitle
\begin{abstract}
Embodied agents face a critical dilemma that end-to-end models lack interpretability and explicit 3D reasoning, while modular systems ignore cross-component interdependencies and synergies. To bridge this gap, we propose the Dynamic 3D Vision-Language-Planning Model (D3D-VLP).
Our model introduces two key innovations: 1) A Dynamic 3D Chain-of-Thought (3D CoT) that unifies planning, grounding, navigation, and question answering within a single 3D-VLM and CoT pipeline; 2) A Synergistic Learning from Fragmented Supervision (SLFS) strategy, which uses a masked autoregressive loss to learn from massive and partially-annotated hybrid data. This allows different CoT components to mutually reinforce and implicitly supervise each other. To this end, we construct a large-scale dataset with 10M hybrid samples from 5K real scans and 20K synthetic scenes that are compatible with online learning methods such as RL and DAgger. Our D3D-VLP achieves state-of-the-art results on multiple benchmarks, including Vision-and-Language Navigation (R2R-CE, REVERIE-CE, NavRAG-CE), Object-goal Navigation (HM3D-OVON), and Task-oriented Sequential Grounding and Navigation (SG3D). Real-world mobile manipulation experiments further validate the effectiveness. The code is available at \href{https://github.com/MrZihan/D3D-VLP}{https://github.com/MrZihan/D3D-VLP}.
\end{abstract}    
\section{Introduction}
\label{sec:intro}

\begin{figure}
\noindent\begin{minipage}[h]{1\columnwidth}%
\begin{center}
\includegraphics[width=1.\columnwidth]{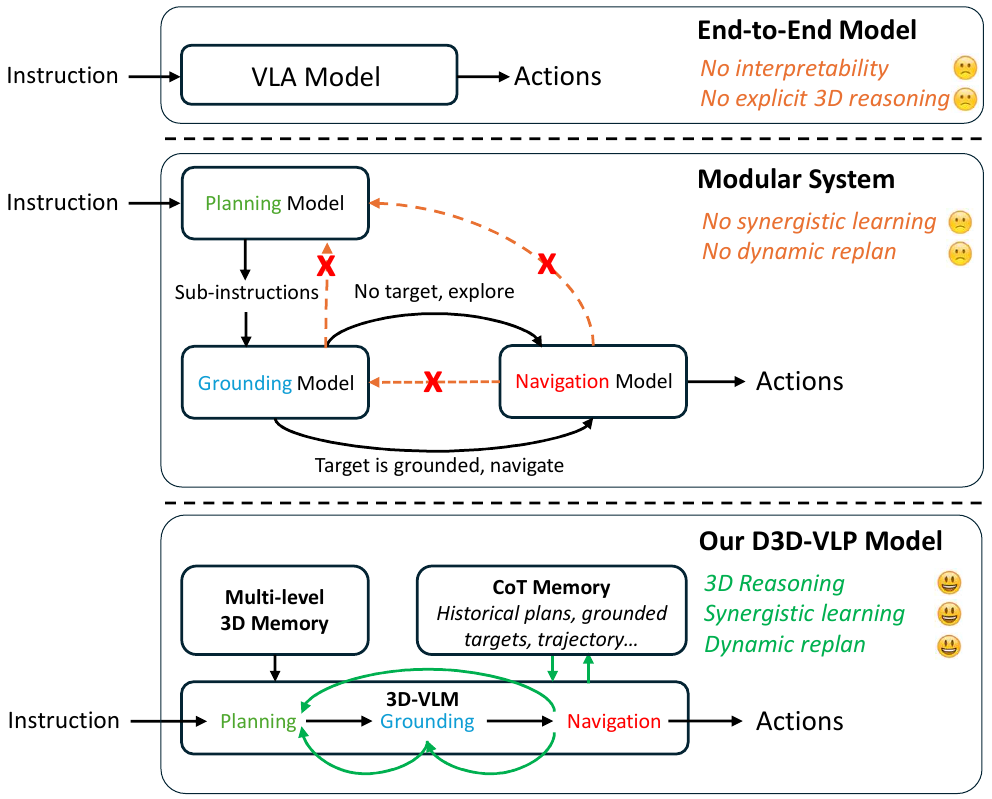}
\par\end{center}%
\end{minipage}
\vspace{-5pt}
\caption{\textbf{Model Architecture Comparison.} The end-to-end models directly map instructions to navigation actions, and modular systems assemble multiple specialized components. Our \OurModel{} employs a single 3D-VLM with 3D CoT to unify planning, grounding, and navigation 
\gh{for} synergistic learning and planning.}
\label{fig:introduction}
\vspace{-6mm}
\end{figure}

Effective 3D vision-language grounding~\cite{chen2020scanrefer,multi3drefer,3dgqa,sg3d} and navigation~\cite{anderson2018vision,krantz2020beyond,rxr,qi2020reverie,wang2025navrag,hm3d-ovon} are critical capabilities for embodied agents
\gh{to achieve exploration of large-scale 3D scenes and} locate task-relevant objects. However, existing methodologies present a fundamental dilemma.
On the one hand, most end-to-end embodied navigation models~\cite{zhang2024navid,zhang2024uninavid,cheng2024navila,wei2025streamvln,wang2025dynam3d} directly output navigation actions, 
\gh{which bypasses} explicit 3D grounding and reasoning processes. 
\gh{However, this simple \textit{black-box} approach}
impairs 
\gh{the} ability to output precise target locations and introduces limitations for tasks 
\gh{that require} long-horizon planning~\cite{sg3d} or handling multiple targets.
On the other hand, specialized 3D grounding models~\cite{chen2024grounded,3d-concept-nerf,mvt3d,vil3dref,pq3d,3d-vista,zhu2024llava} that focus on precise 
\gh{localization} often require complete 
\gh{point cloud of the scene} or comprehensive RGB-D image sequences as input. This 
\gh{dependence} on complete \gh{and} offline information fundamentally limits their deployment in large-scale, unseen, and dynamic real-world environments~\cite{wang2025dynam3d}, where agents must rely on incomplete \gh{and} real-time observations.

\gh{Figure~\ref{fig:introduction} shows some modular systems~\cite{long2024instructnav,chen2025affordances,wang2025dreamnav,liu2024ok,liu2024dynamem} have achieved commendable performance in their effort to bridge the gap by the integration of Large Language Model (LLM) planners, grounding models, and navigation models.}
However, these individual components 
\gh{remain essentially} disjunct \gh{and} operating in a multi-stage pipeline. 
\gh{For example}, the planner often cannot dynamically update its plan based on real-time feedback from the grounding and navigation modules. Furthermore, these modules are frequently trained on disparate datasets, creating a significant domain gap that impedes cohesive operation~\cite{liu2024ok}.

To address these limitations, we propose the \textbf{Dynamic 3D Vision-Language-Planning Model (\OurModel{})}. As shown in Figure~\ref{fig:introduction} (bottom), our model is designed to bridge the gap between interpretable modular systems and high-performance end-to-end models by introducing two key innovations:
\textbf{\textit{1) Dynamic 3D Chain-of-Thought (3D CoT) Pipeline.}} Unlike multi-stage modular systems, \gh{our} \OurModel{} reformulates planning, grounding, and navigation as a single unified autoregressive task within a 3D-VLM~\cite{wang2025dynam3d}. The \textit{dynamic} nature of our 3D CoT is enabled by a \textit{\textbf{CoT Memory}} feedback loop, which feeds historical plans, grounded targets, and 
\gh{trajectory of the agent back into the context of the model. This mechanism makes the agent stateful and aware of its own progress. It can directly address the static nature of prior planners by performing \textit{\textbf{dynamic replanning}} when a plan is blocked or a target is missing.} 
%
\textbf{\textit{2) Synergistic Learning from Fragmented Supervision (SLFS) Strategy.}} \gh{This strategy trains an unified model without requiring millions of perfectly-annotated samples.}
\gh{Specifically, it} allows \gh{our} \OurModel{} to effectively learn from our 10M-sample hybrid dataset \gh{that comprises} 
a small set of \textit{gold} samples and massive quantities of partially-annotated data \gh{such as navigation-only.} 
By using a masked autoregressive loss, the gradient from an available annotation \gh{such as a correct navigation action} 
back-propagates through the shared 3D-VLM \gh{to implicitly supervise and reinforce the internally-generated CoT pipeline of the model.}
This allows all components to mutually supervise and reinforce each other \gh{to achieve \textit{\textbf{synergistic learning}} that is lacking in disjunct modules.}

In summary, our main contributions are:
\begin{itemize}
\item We propose \OurModel{}, a 3D vision-language-planning model that unifies multi-step planning, grounding, and navigation \gh{in unseen and dynamic environments} within a single 3D memory and CoT pipeline. 
\item We introduce the Synergistic Learning from Fragmented Supervision (SLFS) strategy and a supporting 10M-sample large-scale hybrid dataset. This enables our \OurModel{} to learn complex 3D CoT reasoning from massive and partially-annotated sources.
\item We demonstrate new state-of-the-art performance with \gh{our} \OurModel{} across a diverse range of embodied navigation and grounding benchmarks 
\gh{such as} R2R-CE, REVERIE-CE, NavRAG-CE, HM3D-OVON, and SG3D, and validate its effectiveness in real-world mobile manipulation experiments.
\end{itemize}
\section{Related Work}
\label{sec:related_work}


\noindent \textbf{3D Visual Grounding.} Grounding is a core ability for 3D Vision-Language Models (3D-VLMs)~\cite{3d-vista,huang2023embodied,pq3d, lee2024segment, jia2024sceneverse,zhu2024llava,zheng2024video,qi2025gpt4scene}. These approaches vary by input modality, \gh{where} some encode full-scene point clouds~\cite{vil3dref,3d-llm,3d-vista,jia2024sceneverse,pq3d} \gh{while} others process object-level segments~\cite{concept-graph}, or aggregate multi-view 2D features into 3D representations~\cite{zhu2024llava,zheng2024video,zheng2025learning}. These methods are primarily designed for offline analysis 
\gh{with the assumption of an available} complete and static 3D scene. This makes them less directly applicable for embodied agents in unseen, partially observable or dynamic real-world environments.

\smallskip
\noindent \textbf{Embodied Navigation.} Recent works~\cite{zhang2024navid,zhang2024uninavid,cheng2024navila,wei2025streamvln,zheng2024towards,zhou2024navgpt} \gh{show notable performance improvements by using} 
end-to-end large models that directly map instruction and streaming video into navigation actions.
\gh{Although} effective for trajectory following, these approaches often bypass explicit 3D grounding and multi-step planning processes, such grounding-blind design limits their applicability in tasks requiring precise target grounding. 
\gh{Some methods often lack explicit 3D grounding or CoT pipeline~\cite{wei2022chain} for long-horizon planning and multi-target tasks despite their use of 3D-VLMs~\cite{wang2025dynam3d} for navigation.}


\smallskip
\noindent \textbf{Modular Planning Systems.}
To handle long-horizon tasks~\cite{sg3d,song2022one,song2025towards}, many systems adopt a modular architecture~\cite{long2024instructnav,chen2025affordances,wang2025dreamnav,internnav2025} \gh{in which LLMs are often used as high-level planners.} 
These frameworks typically operate in a multi-stage pipeline, \gh{where} an LLM decomposes the task, a grounding module identifies targets, and a navigation policy executes low-level actions. 
\gh{However, a disjoint architecture hampers dynamic plan updates from real-time feedback. It also leaves domain gaps between modules that are trained on disparate datasets.}

\smallskip
\noindent \textbf{Embodied Chain-of-Thought (CoT).}
Recent works explore integrating Chain-of-Thought (CoT)~\cite{wei2022chain} reasoning into embodied agent, 
\gh{but} often without deep 3D grounding and reasoning. NavCoT~\cite{lin2024navcot} proposes a text-based navigational chain-of-thought that 
\gh{hallucinates} the future observation 
\gh{instead of} grounding in \gh{the} actual observed 3D environment.
Embodied CoT~\cite{zawalskirobotic} grounds 2D bounding boxes and reason with 2D visual features, which is tied to the immediate 2D observation 
\gh{instead of} persistent 3D representations. Some models 
\gh{such as} SceneCOT~\cite{linghu2025scenecot} attempt explicit 3D grounding and resemble multiple disparate modules without a unified CoT model.


\section{Our Method}
\label{sec:method}

\begin{figure*}[ht]
\makebox[\textwidth][c]
{\includegraphics[width=0.82\paperwidth]{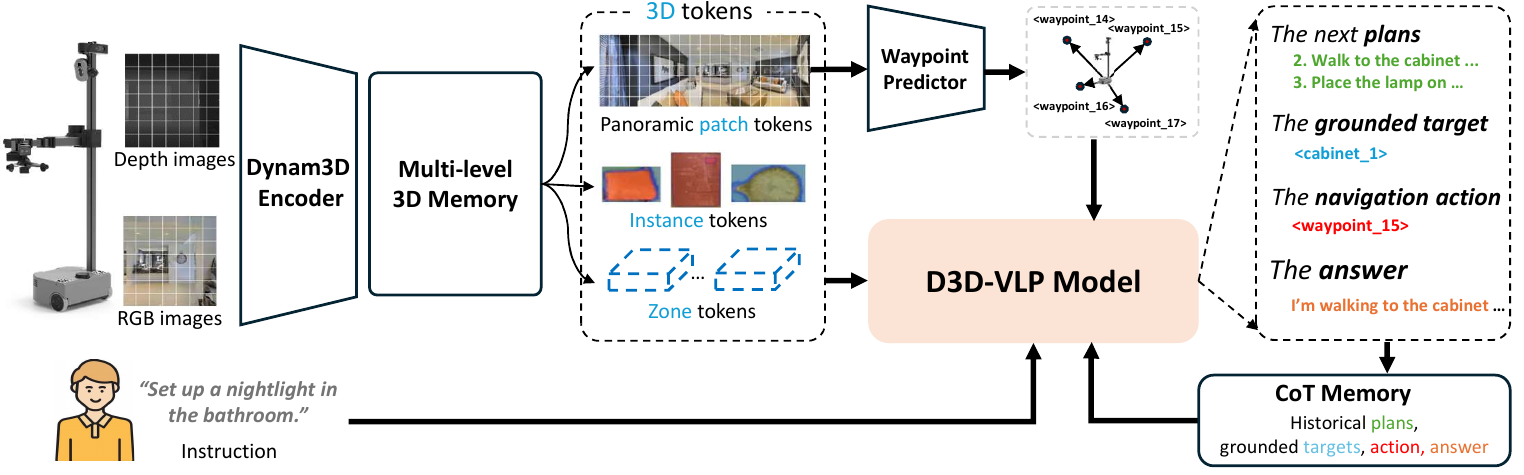}}
\vspace{-20pt}
\caption{\textbf{Framework of our \OurModel{} model.} Given an instruction and streaming posed RGB-D images, a Dynam3D Encoder~\cite{wang2025dynam3d} builds and updates a Multi-level 3D Memory. This memory provides structured 3D tokens (\ie panoramic patch, instance, and zone tokens) to the core \OurModel{} model and a Waypoint Predictor. \gh{Our} \OurModel{} model then integrates these 3D tokens, the instruction, candidate waypoints, and historical context from the CoT Memory to autoregressively generate a unified 3D Chain-of-Thought (CoT) sequence, 
\gh{which includes} the next plans, the grounded target, and the navigation action. Finally, this output updates the CoT Memory 
\gh{to create} a dynamic feedback loop for stateful reasoning and replanning.}
\label{fig:framework}
\vspace{-10pt}
\end{figure*}

\begin{figure*}[ht]
\makebox[\textwidth][c]
{\includegraphics[width=0.8\paperwidth]{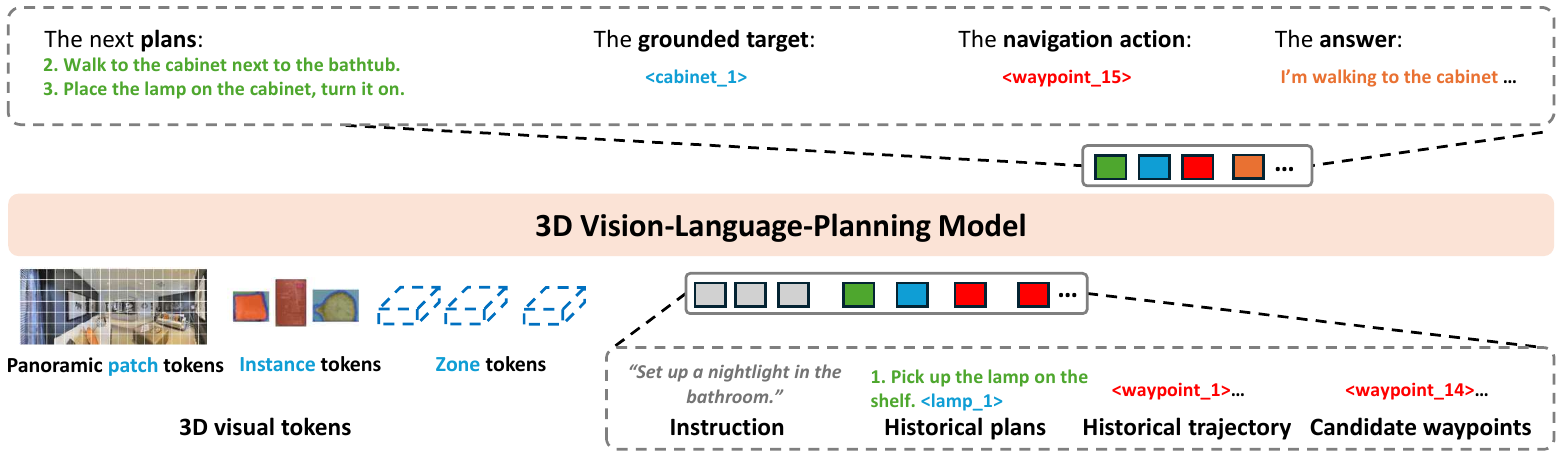}}
\vspace{-15pt}
\caption{\textbf{The Unified Autoregressive Formulation of our \OurModel{} model.} The core 3D Vision-Language-Planning model takes a comprehensive set of inputs: the natural language instruction, multi-level 3D visual tokens (\ie panoramic, instance, and zone) , candidate waypoints, and the historical CoT Memory (including past plans and trajectory). It then autoregressively generates a single \gh{and} unified 3D Chain-of-Thought (CoT) sequence. This multimodal output stream explicitly contains the next plans, the grounded target, the selected navigation action, and a natural language answer.}
\label{fig:3d_cot}
\vspace{-10pt}
\end{figure*}

\noindent \textbf{Overview.} Figure~\ref{fig:framework} shows the framework of our \OurModel{}, a unified 3d vision-language-planning model that integrates planning, grounding, and navigation. At each timestep, 
\gh{we use the encoder of Dynam3D~\cite{wang2025dynam3d} to process}
streaming posed RGB-D images to update a dynamic Multi-level 3D Memory. This memory provides structured 3D tokens to the pre-trained waypoint predictor and the 3D-VLM. The waypoint predictor 
\gh{would} provide several candidate waypoints around the agent. 
\gh{As shown in Figure~\ref{fig:3d_cot}, the 3D-VLM receives the instruction and consults its historical CoT memory. It then autoregressively generates a unified token sequence, including planning text, grounded 3D tokens, the selected next waypoint, and answer text.}
\gh{The updated CoT Memory creates a dynamic feedback loop for replanning and the next decision step.}

\subsection{Multi-level 3D Perception}
\label{sec:memory}
\gh{The 3D CoT is the reasoning core of our \OurModel{}, which requires a persistent and structured representation of the 3D world.}
\gh{To this end, we employ a foundational 3D perception model~\cite{wang2025dynam3d} that processes streaming posed RGB-D images and maintains a multi-level 3D memory. A waypoint predictor then proposes navigable candidate waypoints.}

\smallskip
\noindent \textbf{Dynamic 3D Tokens.} 
\gh{The Dynam3D encoder~\cite{wang2025dynam3d} builds a hierarchical scene representation by first projecting 2D patch features into the 3D space, and then encoding the resulting 3D feature points with rich semantic and geometric information.}
These 3D feature points ($\mathcal{M}_{\text{patch}}$) are then aggregated using transformer-based encoders into object-centric \textbf{Instance Tokens} ($\mathcal{M}_{\text{inst}}$) and coarse-grained spatial \textbf{Zone Tokens} ($\mathcal{M}_{\text{zone}}$). 
\gh{Furthermore, given the current camera pose, we render \textbf{Panoramic Patch Tokens} $(\mathcal{V}_{\text{patch}})$ from 3D feature points $(\mathcal{M}_{\text{patch}})$ using generalizable feature fields~\cite{wang2024g3d}. This provides local fine-grained 3D perception following Dynam3D~\cite{wang2025dynam3d}.}
This structured multi-level 3d representations $\mathcal{M}_{t} = (\mathcal{V}_{\text{patch}}, \mathcal{M}_{\text{inst}}, \mathcal{M}_{\text{zone}})$ provides the rich and hierarchical 3D visual context that our 3D-VLM (\cf Section~\ref{sec:cot}) requires for its unified reasoning process.

\smallskip
\noindent \textbf{Waypoint-based Action Space.} 
\gh{We train a waypoint predictor instead of following prior works in using text-based actions 
\gh{such as} “move forward 0.5m” or “turn left 15 degrees” to simplify the action space and align the spatial semantics between navigation actions and multi-level 3D tokens.}
\gh{Similar to~\cite{Hong2022bridging},} this waypoint predictor inputs panoramic patch tokens ($\mathcal{V}_{\text{patch}}$) and 12 query tokens (spaced 30° apart) into a multi-layer transformer 
to output nearby navigable locations.

\smallskip
\noindent \textbf{Unified 3D Spacial Embedding.} For the multi-level 3D tokens and waypoints, we employ a unified 3D spatial embedding to align their spatial semantics. Specifically, for each timestep $t$, we transform the global coordinates of the 3D tokens into the agent-centric camera coordinate system using the current camera pose 
\gh{to get} $P_t = (P_{\text{patch}}, P_{\text{inst}}, P_{\text{zone}})$. Similarly, we obtain the waypoint coordinates $P_{\text{wayp}}$. For each 3D coordinate point $P_t=(x_t,y_t,z_t)$, the relative distance $D_t$ and the relative horizontal angle $\theta_t$ to the agent can be easily calculated. 
\gh{Subsequently, we get the corresponding spatial embeddings $\mathcal{P}_t$ by feeding $(P_t, D_t, \cos(\theta_t), \sin(\theta_t))$ into an MLP-based $\operatorname{spatial\_encoder(\cdot)}$.}

\subsection{Dynamic 3D Chain-of-Thought (3D CoT)}
\label{sec:cot}
The core of our agent is the 3D-VLM (\cf Figures \ref{fig:framework}-\ref{fig:3d_cot}), which directly addresses the \textit{disjunct} and \gh{the} \textit{no synergistic learning} limitations of modular systems (\cf Figure~\ref{fig:introduction}). This is achieved by reformulating the entire embodied task from high-level task decomposition (planning) and object localization (grounding) to low-level action (navigation) as a single and unified autoregressive generation problem.


\smallskip
\noindent \textbf{{Unified Autoregressive Formulation}.} The 3D-VLM is implemented as an autoregressive multimodal model loaded from a pre-trained
NVILA-Lite-2B~\cite{liu2025nvila}. At each timestep $t$, it takes a comprehensive context as input:

\begin{enumerate}[label=\alph*), itemsep=2pt, topsep=4pt]

\item The \textbf{\textit{Multi-level 3D Representations}} $\mathcal{M}_{t}\oplus\mathcal{P}_{t} = (\mathcal{V}_{\text{patch}}, \mathcal{M}_{\text{inst}}, \mathcal{M}_{\text{zone}})\oplus(\mathcal{P}_{\text{patch}}, \mathcal{P}_{\text{inst}}, \mathcal{P}_{\text{zone}})$. \gh{Similar to Dynam3D~\cite{wang2025dynam3d},} these 3D tokens are projected into the 
\gh{the latent space of the VLM, \ie NVILA-Lite-2B~\cite{liu2025nvila}} with a MLP-based $\operatorname{projector(\cdot)}$ through contrastive learning.  

\item The \textbf{\textit{Instruction}} text tokens $\mathcal{I}$.

\item The historical \textbf{\textit{CoT Memory}} $\mathcal{C}_{t-1}$ from 
the previous timestep, as described \gh{later in Paragraph
\symref{par:cot_memory}{CoT Memory$^\dagger$}.}

\item The candidate waypoints embeddings $\mathcal{P}_{\text{wayp}}$.

\end{enumerate}

The 3D-VLM is then trained to autoregressively generate a single coherent sequence $\mathcal{S}_{t}$ that represents its complete chain-of-thought:
\begin{equation}
p(\mathcal{S}_{t} \mid \mathcal{I}, \mathcal{M}_{t}\oplus\mathcal{P}_{t}, \mathcal{C}_{t-1}).
\end{equation}
As illustrated in Figure~\ref{fig:3d_cot}, this output sequence $\mathcal{S}_{t}$ is a structured concatenation of multimodal tokens $\mathcal{S}_{t} = (\mathcal{T}_{\text{plan}}, \mathcal{T}_{\text{ground}}, \mathcal{T}_{\text{nav}}, \mathcal{T}_{\text{answer}})$:
\begin{enumerate}[label=\alph*), itemsep=2pt, topsep=4pt]
    \item $\mathcal{T}_{\text{plan}}$: A sequence of natural language tokens representing the next step(s) of the high-level plan, \eg 
    “The next plans: 2. Walk to the cabinet next to the bathtub. 3. Place the lamp on the cabinet, turn it on”.
    
    \item $\mathcal{T}_{\text{ground}}$: This sequence is not only language generation. Instead, the model outputs multimodal tokens: “The grounded: target\texttt{<target\_1>}, target\texttt{<target\_2>}”. These special tokens (\eg \texttt{<target\_1>}) are autoregressively followed by the word “target” and points to the corresponding 3D tokens from the $\mathcal{M}_{\text{t}}$ (\eg \texttt{<cabinet\_1>}) in Figure~\ref{fig:3d_cot}. This forces the model to make an explicit \gh{and} interpretable grounding decision within the CoT sequence.
    
    \item $\mathcal{T}_{\text{nav}}$: The model generates the navigation sequence “The navigation action: waypoint \texttt{<waypoint>}”, The special token \texttt{<waypoint>} is followed by the word “waypoint” and selects a waypoint from the candidate waypoints. The agent can move to this selected waypoint through \gh{the} action controller.
    
    \item $\mathcal{T}_{\text{answer}}$: A sequence of language tokens for internal monologue, question answering, or status description.
\end{enumerate}
This formulation is not a simple chain-of-text. It is a multimodal chain-of-thought, where language reasoning ($\mathcal{T}_{\text{plan}}$) is directly followed by an explicit 3D grounding action ($\mathcal{T}_{\text{ground}}$) and a navigation action ($\mathcal{T}_{\text{nav}}$) 
within one autoregressive pass.

\smallskip
\noindent \textbf{Grounding within 3D Tokens.} 
\gh{The 3D-VLM first outputs a latent \texttt{<target>} token, which is passed through an MLP-based $\operatorname{grounding\_head}(\cdot)$. We then compute dot-product similarities between the resulting feature and all 3D tokens from $\operatorname{projector}(\mathcal{M}_{t}\oplus\mathcal{P}_{t})$, and a special
\texttt{<grounding\_none>} token.}
The token with the highest similarity is selected as the grounded target and input into 3D-VLM for  the subsequent autoregressive process.

\smallskip
\noindent \textbf{Navigation within Candidate Waypoints.} 
\gh{Similar to the grounding process, the latent \texttt{<waypoint>} token is first passed through an MLP-based $\operatorname{navigation\_head}(\cdot)$. We then compute dot-product similarities between the resulting feature and all candidate waypoint embeddings from $\operatorname{projector}(\mathcal{P}_{\text{wayp}})$, and a special \texttt{<navigation\_stop>} token.} 
The candidate waypoint embedding with the highest similarity is selected as the next movement target, 
\gh{with \texttt{<stop>} indicating} that the destination has been reached.
%
\gh{Note that the model does not output \texttt{<stop>} 
for the sub-goals within the sub-instruction from planning.}
Instead, after selecting the optimal next \texttt{<waypoint>}, it outputs the text tokens “reached the subgoal” to identify the completion progress of the sub-instruction.

\smallskip
\noindent \textbf{CoT Memory Feedback for Dynamic Replanning$^\dagger$.}
\label{par:cot_memory}
The \textit{dynamic} nature of the 3D CoT is enabled by the memory feedback loop shown in Figures~\ref{fig:framework} and~\ref{fig:3d_cot}. 
\gh{Instead of discarding the generated sequence $\mathcal{S}_{t}$, we parse its components to update the historical CoT Memory for the next timestep:}
\begin{equation}
\mathcal{C}_{t} = \text{Concat}(\mathcal{C}_{t-1}, \text{Parse}(\mathcal{S}_{t})),
\end{equation}
\gh{where} $\text{Parse}(\mathcal{S}_{t}) = (\text{Parse}(\mathcal{T}_{\text{plan}}), \text{Parse}(\mathcal{T}_{\text{ground}}), \text{Parse}(\mathcal{T}_{\text{nav}}))$.
Specifically, Parse($\mathcal{T}_{\text{plan}}$) records the completed sub-instruction, Parse($\mathcal{T}_{\text{ground}}$) records the grounded target token and position, and $\text{Parse}(\mathcal{T}_{\text{nav}})$ records the previously traversed waypoint position. $\mathcal{C}_{t}$ is fed back into the 3D-VLM for the next decision-making step, as shown in Figure~\ref{fig:3d_cot}.

\gh{This mechanism makes the agent explicitly stateful and enables online re-planning, which directly addresses the ``no dynamic replan'' failure mode of static and multi-stage modular systems (\cf Figure~\ref{fig:introduction}). At each step, our \OurModel{} is forced to re-read its own history of previous plans, grounded targets, and executed trajectories in the input context.
When a sub-instruction cannot be satisfied, \eg when the target cannot
be grounded, navigation becomes blocked, or the current plan stalls, the 3D-VLM can interpret these failure signals in $\mathcal{C}_{t}$ and
autoregressively propose a revised plan $\mathcal{S}_{t+1}$. This closed-loop design turns planning into a self-correcting process instead of a one-shot prediction.
}

\subsection{Synergistic Fragmented Supervision}
\label{sec:slfs}
Training our \OurModel{} model with 3D CoT to achieve unified planning, grounding, and navigation remains a highly challenging task due to 
data scarcity and training difficulty.
\gh{To address these challenges, we introduce \emph{\textbf{Synergistic Learning from Fragmented Supervision}} (SLFS), which leverages a large-scale but fragmented 3D CoT dataset together with a masked autoregressive objective such that each partially annotated sample still contributes to joint learning of planning, grounding, and navigation.}


\begin{table}[h]
\caption{Composition of sample annotations in our constructed 3D CoT dataset. The fully annotated gold data is about 175K, and the partially annotated data is about 9.9M.}
\vspace{-5pt}
\centering
\renewcommand{\arraystretch}{0.9}
\resizebox{\columnwidth}{!}{%
  \begin{tabular}{c c c c c} 
    \hline
    \textbf{Data Type} & \textbf{Planning} & \textbf{Grounding} & \textbf{Navigation} & \textbf{\# Samples} \\
    \hline
    1 & \checkmark & \checkmark & \checkmark & 175K \\
    2 & \checkmark & \checkmark & $\times$   & 161K \\
    3 & \checkmark & $\times$   & \checkmark & 14K  \\
    4 & $\times$   & \checkmark & \checkmark & 5.8M \\
    5 & $\times$   & \checkmark & $\times$   & 2.3M \\
    6 & $\times$   & $\times$   & \checkmark & 1.6M \\
    \hline
  \end{tabular}%
}
\label{tab:dataset_stats}
\end{table}

\smallskip
\noindent \textbf{3D CoT Dataset.}
\gh{
Collecting a massive dataset (\eg 10M+ samples) with high-quality
ground-truth for all 3D-CoT components $(\mathcal{T}_{\text{plan}}, \mathcal{T}_{\text{ground}}, \mathcal{T}_{\text{nav}})$ over every sample is economically infeasible. Instead, we construct a large-scale hybrid dataset of 10 million samples by aggregating data from multiple heterogeneous sources.
As summarized in Table~\ref{tab:dataset_stats}, this dataset is fragmented. It contains a small subset of fully annotated \textit{gold} samples
(\eg 175K), and large amounts of partially annotated data
such as 1.6 million samples that only provide navigation annotations.
}

The \textbf{\textit{planning annotations}} (data type 1,2,3) are mainly from SG3D~\cite{sg3d}, Grounded 3D-LLM~\cite{chen2024grounded} and VLN-Trans~\cite{zhang2023vln}. For 3D grounding (data type 5), we primarily collect 3D instance-text pairs from SceneVerse~\cite{jia2024sceneverse}, MMScan~\cite{mmscan}, PQ3D~\cite{pq3d}, and Grounded 3D-LLM~\cite{chen2024grounded}. 
 
\gh{For \textbf{\textit{open-vocabulary object grounding and navigation}} (data type 4), we use annotations in HM3D~\cite{ramakrishnan2habitat,hm3d-sem} and
MP3D~\cite{chang2017matterport3d} scenes from the above grounding datasets that support the Habitat simulator~\cite{habitat,puig2023habitat3}. We further integrate data from HSSD (Habitat)~\cite{khanna2024habitat}, ProcTHOR-10K (Habitat)~\cite{khanna2024habitat,procthor}, and ProcTHOR-Objaverse (AI2-THOR)~\cite{ehsani2024spoc,kolve2017ai2}.}
For \textbf{\textit{vision-and-language navigation}} (data type 6), we mainly adapt data from R2R-CE~\cite{krantz2020beyond}, REVERIE-CE~\cite{qi2020reverie,wang2025dynam3d}, SRDF~\cite{wangbootstrapping}, and NavRAG~\cite{wang2025navrag} to the Habitat simulator~\cite{habitat,puig2023habitat3}.

The 3D scenes for above annotations include posed RGB-D videos from real environments such as ScanNet~\cite{dai2017scannet}, 3RScan~\cite{3rscan}, and ARKitScenes~\cite{baruch1arkitscenes}; real scans such as HM3D~\cite{ramakrishnan2habitat} and MP3D~\cite{chang2017matterport3d}; as well as synthetic scenes including HSSD~\cite{khanna2024habitat}, ProcTHOR-10K~\cite{procthor}, and ProcTHOR-Objaverse~\cite{spoc}.


\medskip
\noindent \textbf{Masked Autoregressive Loss.} The SLFS mechanism is implemented at the loss-computation level. The \OurModel{} always generates the full sequence prediction $\mathcal{S}_{\text{pred}} = (\mathcal{T}_{\text{plan}}^{\text{pred}}, \mathcal{T}_{\text{ground}}^{\text{pred}}, \mathcal{T}_{\text{nav}}^{\text{pred}}, \mathcal{T}_{\text{answer}}^{\text{pred}})$ for every sample in a batch. If the annotation for a CoT component is available, the corresponding multimodal tokens from the annotation are directly concatenated to the existing input tokens. If no annotation is available, the tokens for that CoT component are generated autoregressively, and these tokens are masked out during the cross-entropy loss computation. 

For each sample $i$ in the batch, we construct a ground-truth target sequence $\mathcal{S}_{\text{gt},i}$. Any missing annotation token (\eg $\mathcal{T}_{\text{plan}}$ for a navigation-only sample) is populated with \gh{a} special \texttt{<mask>} token in $\mathcal{S}_{\text{gt},i}$.

The total loss $\mathcal{L}_{CoT}$ is then computed using a standard autoregressive cross-entropy loss, which is masked to ignore the \texttt{<mask>} tokens:
\begin{equation}
\label{eq:loss}
\mathcal{L}_{CoT} = \sum_{i \in \text{Batch}} \sum_{k \in \text{CoT}} \mathcal{H}_{i,k} \cdot \mathcal{L}_{k}(\mathcal{S}_{\text{pred},i}, \mathcal{S}_{\text{gt},i})
\end{equation}
where $\mathcal{H}_{i,k}$ is an indicator mask that is 1 if sample $i$ has a valid annotation for CoT component $k$, and 0 otherwise.

\gh{
This strategy enables synergistic learning across different data types:

\smallskip
\noindent \textbf{\textit{Gold Data.}} These are the fully annotated samples, where all masks ($\mathcal{H}_{i,\text{plan}}, \mathcal{H}_{i,\text{ground}}, \dots$) are set to 1. The model receives explicit end-to-end supervision over the entire CoT sequence. These samples act as \textit{semantic anchors} to guide the model into aligning its internal representations with the correct language, grounding, and action tokens, \eg ``Go to the kitchen and take the bread to the living room table.''.
    
\smallskip
\noindent \textbf{\textit{Partially Annotated Data.}}
Using navigation-only data as an example, where only
$\mathcal{H}_{i,\text{nav}} = 1$. The loss $\mathcal{L}_{\text{nav}}$ is computed solely on
$\mathcal{T}_{\text{nav}}^{\text{pred}}$. However, during the autoregressive forward pass, $\mathcal{T}_{\text{nav}}^{\text{pred}}$ is \emph{conditioned on} the internally generated $\mathcal{T}_{\text{plan}}^{\text{pred}}$
and $\mathcal{T}_{\text{ground}}^{\text{pred}}$ of the model.
As a result, the gradient from $\mathcal{L}_{\text{nav}}$ back-propagates through the shared 3D-VLM and implicitly supervises the planning and grounding modules.
}


\section{Experiments}
\label{sec:experiments}

\begin{table*}[ht]
\caption{Evaluation of embodied navigation benchmarks with monocular camera, $*$ denotes zero-shot method.}
\vspace{-6pt}
\centering
\renewcommand{\arraystretch}{0.9}
\resizebox{\textwidth}{!}{%
\begin{tabular}{l|l|cccc|cccc|cccc|cc}
\hline

\multirow{2}{*}{Methods} & \multirow{2}{*}{System Type} & \multicolumn{4}{c|}{R2R-CE} & \multicolumn{4}{c|}{REVERIE-CE} & \multicolumn{4}{c|}{NavRAG-CE} & \multicolumn{2}{c}{HM3D-OVON} \tabularnewline
\cline{3-16}
& & NE\textdownarrow{} & OSR\textuparrow{} & SR\textuparrow{} & SPL\textuparrow{} & NE\textdownarrow{} & OSR\textuparrow{} & SR\textuparrow{} & SPL\textuparrow{} & NE\textdownarrow{} & OSR\textuparrow{} & SR\textuparrow{} & SPL\textuparrow{} & SR\textuparrow{} & SPL\textuparrow{} \tabularnewline
\hline

CM$^{2}$~\cite{georgakis2022cross} & E2E & 7.02 & 41.5 & 34.3 & 27.6 & - & - & - & - & - & - & - & - & - & - \tabularnewline

WS-MGMap~\cite{chen2022weakly} & E2E & 6.28 & 47.6 & 38.9 & 34.3 & - & - & - & - & - & - & - & - & - & - \tabularnewline

InstructNav$^{*}$~\cite{long2024instructnav} & Modular w/ CoT & 6.89 & - & 31.0 & 24.0 & 7.44 & 31.5 & 25.2 & 19.1 & 9.83 & 24.1 & 17.4 & 10.9 & - & - \tabularnewline

CA-Nav$^{*}$~\cite{chen2025constraint} & Modular & 7.58 & 48.0 & 25.3 & 10.8 & - & - & - & - & - & - & - & - & - & - \tabularnewline

AO-Planner$^{*}$~\cite{chen2025affordances} & Modular w/ CoT & 6.95 & 38.3 & 25.5 & 16.6 & - & - & - & - & - & - & - & - & - & - \tabularnewline

DreamNav$^{*}$~\cite{wang2025dreamnav} & Modular w/ CoT & 7.06 & 41.0 & 32.8 & 29.0 & - & - & - & - & - & - & - & - & - & - \tabularnewline

\textcolor{black}{VLFM$^{*}$~\cite{yokoyama2024vlfm}} & Modular & - & - & - & - & - & - & - & - & - & - & - & - & 35.2 & 19.6 \tabularnewline

\textcolor{black}{DAgRL+OD$^{*}$~\cite{hm3d-ovon}} & Modular & - & - & - & - & - & - & - & - & - & - & - & - & 37.1 & 19.8 \tabularnewline

\textcolor{black}{MTU3D~\cite{zhu2025move}} & E2E & - & - & - & - & - & - & - & - & - & - & - & - & 40.8 & 12.1 \tabularnewline

\textcolor{black}{VLN-3DFF~\cite{wang2024simtoreal}} & E2E & 5.95 & 55.8 & 44.9 & 30.4 & - & - & - & - & - & - & - & - & - & - \tabularnewline

\textcolor{black}{g3D-LF~\cite{wang2024g3d}} & E2E & 5.70 & 59.5 & 47.2 & 34.6 & 6.50 & 41.6 & 34.4 & 23.8 & 8.85 & 31.8 & 21.4 & 13.5 & - & - \tabularnewline

NaVid~\cite{zhang2024navid} & E2E & 5.47 & 49.1 & 37.4 & 35.9 & 6.74 & 36.3 & 26.6 & 20.8 & 9.35 & 29.6 & 19.4 & 13.9 & - & - \tabularnewline

MapNav~\cite{zhang2025novel} & Modular w/ CoT & 4.93 & 53.0 & 39.7 & 37.2 & - & - & - & - & - & - & - & - & - & - \tabularnewline

\textcolor{black}{Uni-NaVid~\cite{zhang2024uninavid}} & E2E & 5.58 & 53.3 & 47.0 & 42.7 & - & - & - & - & - & - & - & - & 39.5 & 19.8 \tabularnewline

\textcolor{black}{NaVILA~\cite{cheng2024navila}} & E2E & 5.22 & 62.5 & 54.0 & 49.0 & - & - & - & - & - & - & - & - & - & - \tabularnewline

\textcolor{black}{Aux-Think~\cite{wang2025think}} & E2E & 5.88 & 54.9 & 49.7 & 41.7 & - & - & - & - & - & - & - & - & - & - \tabularnewline

\textcolor{black}{Dynam3D~\cite{wang2025dynam3d}} & E2E & 5.34 & 62.1 & 52.9 & 45.7 & 6.22 & 48.9 & 40.1 & 28.5 & 8.12 & 38.4 & 24.7 & 18.8 & 42.7 & 22.4 \tabularnewline

\textcolor{black}{MonoDream~\cite{wang2025monodream}} & E2E & 5.45 & 61.5 & 55.8 & 49.1 & - & - & - & - & - & - & - & - & - & - \tabularnewline

\textcolor{black}{StreamVLN~\cite{wei2025streamvln}} & E2E & 4.98 & 64.2 & 56.9 & 51.9 & - & - & - & - & - & - & - & - & - & - \tabularnewline

\textcolor{black}{NavFoM (S.RGB)~\cite{zhang2025embodied}} & E2E & 5.01 & 64.9 & 56.2 & 51.2 & - & - & - & - & - & - & - & - & 43.6 & \textbf{31.3} \tabularnewline

\textcolor{black}{InternVLA-N1~\cite{internnav2025}} & Modular & 4.83 & 63.3 & 58.2 & 54.0 & - & - & - & - & - & - & - & - & - & - \tabularnewline

\hline

\textcolor{black}{\OurModel{} (Ours)} & E2E w/ CoT & \textbf{4.73} & \textbf{67.2} & \textbf{61.3} & \textbf{56.1} & \textbf{5.36} & \textbf{56.9} & \textbf{47.5} & \textbf{34.7} & \textbf{7.57} & \textbf{45.3} & \textbf{31.1} & \textbf{23.9} & \textbf{47.3} & 30.4 \tabularnewline

\hline
\end{tabular}%
}
\label{tab:navigation_sota}
\end{table*}

\begin{table}[ht]
\caption{Evaluation of task-oriented sequential grounding and navigation task on SG3D-Nav~\cite{sg3d} benchmark.}
\vspace{-6pt}
\centering
\resizebox{\columnwidth}{!}{%
\begin{tabular}{l|l|ccc|cc}
\hline
\multirow{2}{*}{Methods} & \multirow{2}{*}{System Type} &
\multicolumn{3}{c|}{Navigation} & \multicolumn{2}{c}{Grounding} \tabularnewline
\cline{3-7}
& & s-SR & t-SR & SPL & s-ACC & t-ACC \tabularnewline
\hline
VideoAgent~\cite{embodied-video-agent,sg3d} & Modular & 14.7 & 3.8 & 10.2 & - & -  \tabularnewline
SenseAct-NN~\cite{khanna2024goat,sg3d}      & E2E     & 12.1 & 7.7 & 10.1 & - & - \tabularnewline
MTU3D~\cite{zhu2025move}                    & E2E     & 23.8 & 8.0 & 16.5 & - & - \tabularnewline
Dynam3D-VisTA~\cite{wang2025dynam3d,3d-vista}& Modular& 26.4 & 9.3 & 15.4 & 21.4 & 4.2 \tabularnewline
\OurModel{} (Ours)                          & E2E w/ CoT &
\textbf{33.7} & \textbf{13.8} & \textbf{21.6} & \textbf{28.3} & \textbf{9.3} \tabularnewline
\hline
\end{tabular}%
}
\label{tab:planning_sota}
\end{table}


\subsection{Experimental Setup}
\label{sec:setup}

\noindent\textbf{Benchmarks.} We evaluate \OurModel{} on a diverse suite of five challenging benchmarks:

\smallskip
\noindent \textbf{1-3) \textit{Vision-and-Language Navigation} (VLN).}
We use \textbf{R2R-CE}~\cite{anderson2018vision,krantz2020beyond}, \textbf{REVERIE-CE}~\cite{qi2020reverie,wang2025dynam3d}, and \textbf{NavRAG-CE}~\cite{wang2025navrag,wang2025dynam3d}. These tasks evaluate the ability \gh{of the agent} to follow natural language instructions 
\gh{that ranges} from step-by-step directions (R2R-CE) to coarse-grained destination descriptions (REVERIE-CE) and complex user-demand instructions (NavRAG-CE).

\smallskip
\noindent \textbf{4) \textit{Object-Goal Navigation} (OVON).} We use \textbf{HM3D-OVON} \cite{hm3d-ovon} to assess the ability \gh{of the agent} in navigating to open-vocabulary object categories in unseen large-scale environments.

\smallskip
\noindent \textbf{5) \textit{Task-Oriented Sequential Grounding and Navigation} (SG3D).} We utilize the \textbf{SG3D} \cite{sg3d} benchmark, a highly challenging task requiring the agent to interpret a multi-step plan (\eg “Make coffee”) and sequentially navigate to and ground context-dependent targets (\eg “find a cup”, “go back to the other table”). 
\gh{Our setting is more challenging than the original benchmark which uses full scene since we require online grounding during navigation.}
Due to the 
\gh{infeasibility of exactly reproducing the ground-truth plan}, we follow the original SG3D setting during evaluation that provides the ground-truth planning text as model input.



\medskip
\noindent\textbf{Evaluation Metrics.} 
\gh{We evaluate our model using standard navigation metrics and stricter sequential grounding metrics as follows:}

\smallskip
\noindent \textbf{\textit{Navigation Benchmarks.}} For R2R-CE, REVERIE-CE, NavRAG-CE, HM3D-OVON, and SG3D Navigation, we report the standard metrics: Navigation Error (NE), Success Rate (SR), Oracle Success Rate (OSR), Success weighted by Path Length (SPL), step-Navigation Success Rate (s-SR), and task-Navigation Success Rate (s-SR).

\smallskip
\noindent \textbf{\textit{Grounding Benchmarks.}}
For the SG3D Grounding task, we introduce two stringent metrics:
\begin{enumerate}[label=\roman*., leftmargin=15pt]

        \smallskip
        \item \textbf{s-ACC (step-Accuracy):} A single step is successful \textit{only if} the agent successfully navigates to the correct target \textit{and} correctly grounds the target, \ie the point in the ground-truth instance point cloud that is closest to the grounded 3D token belongs to the target instance. This is much more challenging than the standard navigation-only step-SR.

        \smallskip
        \item \textbf{t-ACC (task-Accuracy):} A task is successful \textit{only if} all navigation and grounding actions for \textit{every step} in the sequence are correct.
\end{enumerate}

 

\medskip
\noindent\textbf{Implementation Details.} \gh{We next outline the key architectural choices and training setup of our \OurModel{}:}

\smallskip
\noindent \textbf{\textit{Model Architecture.}} \gh{Our} \OurModel{} utilizes the Dynam3D Encoder~\cite{wang2025dynam3d} as the 3D perception backbone. This module processes streaming posed RGB-D images to maintain the multi-level 3D memory \gh{that comprises} patch-level feature field~\cite{wang2024g3d}, instance tokens $\mathcal{M}_{inst}$, zone tokens $\mathcal{M}_{zone}$, \etc. The core reasoning module is a 3D-VLM based on a pre-trained NVILA-Lite-2B model~\cite{liu2025nvila}. The action space is defined by a waypoint predictor, which generates navigable candidates from panoramic patch tokens.

\smallskip
\noindent \textbf{\textit{Training.}} The model is trained using our Synergistic Learning from Fragmented Supervision (SLFS) strategy (\cf Section~\ref{sec:slfs}) with our large-scale hybrid dataset of 10M samples in Table~\ref{tab:dataset_stats}. The model is trained with the masked autoregressive cross-entropy loss  for 100K episodes ($\sim$14 days) on 4 RTX 6000 Ada GPUs with total batch size 8. Due to \textit{limited computational resources}, we cannot use all available data for training. As shown in Table~\ref{tab:dataset_stats}, we sample data of types 1, 2, and 3; types 4 and 5; and type 6 with approximately a 1:1:1 probability during training. All samples within the same batch are kept consistent in terms of data type, and the samples containing navigation tasks are trained online on the simulators~\cite{habitat,kolve2017ai2} with a DAgger augmentation strategy~\cite{chen2022think,an2024etpnav}.


\subsection{Comparison with State-of-the-Art Methods}
\label{sec:sota}

\gh{Table~\ref{tab:navigation_sota} shows the comparison results of our \OurModel{} with other existing methods on various embodied navigation benchmarks.}

\smallskip
\noindent\textbf{VLN Performance (R2R-CE, REVERIE-CE, NavRAG-CE).} \gh{Our} \OurModel{} establishes a new state-of-the-art across all end-to-end (E2E) models and modular systems on three VLN benchmarks. On the widely-used R2R-CE benchmark, \gh{our} \OurModel{} sets a new SOTA with 61.3\% SR and 56.1\% SPL. \gh{This is a substantial improvement over prior end-to-end models, including StreamVLN~\cite{wei2025streamvln} (+4.4\% SR, +4.2\% SPL) and
NavFoM~\cite{zhang2025embodied} (+5.1\% SR, +4.9\% SPL), and also surpasses the strongest modular system, InternVLA-N1~\cite{internnav2025} (+3.1\% SR, +2.1\% SPL).}

\gh{A critical comparison is against the perception baseline Dynam3D~\cite{wang2025dynam3d}. The sophisticated reasoning architecture of our \OurModel{}}
provides a substantial boost of +8.4\% SR and +10.4\% SPL over the Dynam3D baseline. \gh{This result strongly suggests that our 3D CoT and unified architecture are the main contributors to the SOTA performance, and not merely from the strong 3D perception backbone.}

\smallskip
\noindent\textbf{ObjectNav Performance (HM3D-OVON).} On the challenging open-vocabulary object navigation task, \gh{our} \OurModel{} again achieves a new SOTA result with 47.3\% SR and 30.4\% SPL. This outperforms both the strongest E2E model, NavFoM (43.6\% SR, 31.3\% SPL), and our baseline method, Dynam3D (42.7\% SR, 22.4\% SPL). 
\gh{This demonstrates that our unified grounding-and-navigation pipeline is highly effective for object-centric tasks, and not only for instruction-following VLN.}

\subsection{
Long-Horizon Grounding and Planning}
\label{sec:long_horizon}

The SG3D benchmark is specifically designed to evaluate 
planning, grounding, and memory capabilities in long-horizon stateful tasks \gh{of an agent}.

\gh{Table~\ref{tab:planning_sota} shows that our} \OurModel{} demonstrates clear superiority on this complex long-horizon benchmark. \gh{Particularly,} our model achieves a step-level navigation SR of 33.7\% \gh{that} significantly 
\gh{outperforms} the next-best modular baseline (Dynam3D-VisTA, 26.4\% SR) and the SOTA E2E model (MTU3D, 23.8\% SR).

The true challenge of SG3D lies in sequential consistency measured by task-level metrics (t-SR and t-ACC). The baselines show a massive drop-off from step-success to task-success. For example, the Dynam3D-VisTA modular baseline, which pairs the strong 3D perception and navigation baseline model~\cite{wang2025dynam3d} with a 3D grounding model~\cite{3d-vista} achieves a 21.4\% s-ACC but only a 4.2\% t-ACC. It means that although \textit{$\sim$21\% individual steps} are correctly executed, only \textit{$\sim$4\% full tasks} are completed. This confirms the hypothesis that modular memory-less systems cannot handle the contextual dependencies \gh{such as “go back to it”} 
inherent in sequential planning.

In contrast, \gh{our} \OurModel{} achieves a t-ACC of 9.3\%, which is a \textbf{121\% relative improvement} over the Dynam3D-VisTA baseline (4.2\%). Similarly, our navigation t-SR of 13.8\% shows a \textbf{72.5\% relative improvement} over the SOTA MTU3D (8.0\%)  and a 48\% improvement over Dynam3D-VisTA (9.3\%).

This 
\gh{significant} improvement in task-level metrics is direct evidence that our \textbf{Dynamic 3D CoT} (\cf Section~\ref{sec:cot})  and \textbf{CoT Memory} (\cf Section~\ref{sec:memory}) are effective. By feeding historical plans, grounded targets, and trajectories back into the VLM, our \OurModel{} maintains state, resolves temporal ambiguities, and possesses replanning capabilities that are absent in both E2E and static modular systems.

\subsection{Ablation Studies}
\label{sec:ablation}

\begin{table}[ht]
\caption{Ablation study on components and training data.}
\vspace{-7pt}
\centering
\resizebox{\columnwidth}{!}{%
\begin{tabular}{l|l|ccc|cc}
\hline
\multirow{2}{*}{Settings} & \multirow{2}{*}{Training data} &
\multicolumn{3}{c|}{R2R-CE Nav.} & \multicolumn{2}{c}{SG3D Grounding} \tabularnewline
\cline{3-7}
& & OSR & SR & SPL & s-ACC & t-ACC \tabularnewline
\hline
All pipeline & All data & 67.2 & \textbf{61.3} & \textbf{56.1} & \textbf{28.3} & \textbf{9.3} \tabularnewline
All pipeline & w/ $\mathcal{T}_{\text{plan}}$\footnotesize{(Type 1,2,3 in Tab.~\ref{tab:dataset_stats})} &
55.6 & 46.2 & 38.7 & 22.5 & 5.5 \tabularnewline
All pipeline & w/o $\mathcal{T}_{\text{plan}}$\footnotesize{(Type 4,5,6 in Tab.~\ref{tab:dataset_stats})} &
\textbf{67.8} & 60.8 & 55.4 & 24.3 & 5.3 \tabularnewline
\footnotesize{w/o CoT Memory} & All data &
65.7 & 56.5 & 48.7 & 19.4 & 4.1 \tabularnewline
\footnotesize{Text-based actions} & All data &
66.3 & 56.4 & 48.8 & 27.8 & 8.6 \tabularnewline
\hline
\end{tabular}%
}
\label{tab:ablation}
\vspace{-6pt}
\end{table}

To validate the design of \gh{our} \OurModel{}, we conduct a thorough ablation study \gh{in} Table~\ref{tab:ablation} on the R2R-CE and SG3D benchmarks. We \gh{validate and} analyze the \textbf{\textit{three core contributions}} of our work: 

\gh{
\begin{enumerate}[label=\arabic*., leftmargin=*, itemsep=4pt, topsep=4pt]
\item \textbf{Synergistic Learning (SLFS) and Training Data.}
Rows 2 and 3 of Table~\ref{tab:ablation} analyze our SLFS strategy.
Training \emph{only} on samples with planning annotations (w/ $\mathcal{T}_{\text{plan}}$, types 1--3 in Table~\ref{tab:dataset_stats}) leads to a large performance drop (SR 61.3\% $\rightarrow$ 46.2\% on R2R-CE; t-ACC 9.3\% $\rightarrow$ 5.5\% on SG3D), which confirms that high-quality fully annotated data alone is too scarce to train a robust agent.

The ablation also reveals two complementary roles of SLFS:
1) SLFS enables the model to exploit massive partially annotated data (w/o $\mathcal{T}_{\text{plan}}$, types 4--6) to recover strong navigation performance (60.8\% SR on R2R-CE, close to 61.3\% with all data).
2) The small set of gold-standard planning data (w/ $\mathcal{T}_{\text{plan}}$) remains crucial for complex planning, where SLFS effectively leverages it to boost SG3D t-ACC from 5.3\% to 9.3\%.

\item \textbf{Dynamic 3D CoT (CoT Memory).} 
Row 4 (w/o CoT Memory) of Table~\ref{tab:ablation} ablates the core of our Dynamic 3D CoT, which feeds historical plans, groundings, and trajectories back into the VLM.
Removing this memory feedback loop causes a clear performance drop on the standard R2R-CE task (SR 61.3\% $\rightarrow$ 56.5\%).
The SG3D Grounding results further provide the strongest evidence for our central claim that CoT Memory is the critical mechanism for long-horizon sequential tasks.
Without it, the agent degenerates from a planning and stateful controller into a reactive and memory-less one, and the task-level accuracy t-ACC collapses from 9.3\% to 4.1\%.

\item \textbf{Unified 3D Spatial Embedding for Waypoint-based Action Space.} Row 5 of Table~\ref{tab:ablation} shows replacing the waypoint-based action space with direct text-based actions such as ``turn left 30 degrees'' and ``move forward 1.5 meters'' leads to a substantial drop in navigation performance on R2R-CE. This suggests that the waypoint-based action space exploits the 3D reasoning capabilities of our \OurModel{} more effectively by operating within a unified
3D spatial embedding space from Section~\ref{sec:memory}.

\end{enumerate}
}

\subsection{Real-World Mobile Manipulation Experiments}

\begin{figure}
\vspace{-8pt}
\noindent\begin{minipage}[h]{1\columnwidth}%
\begin{center}
\includegraphics[width=0.8\columnwidth]{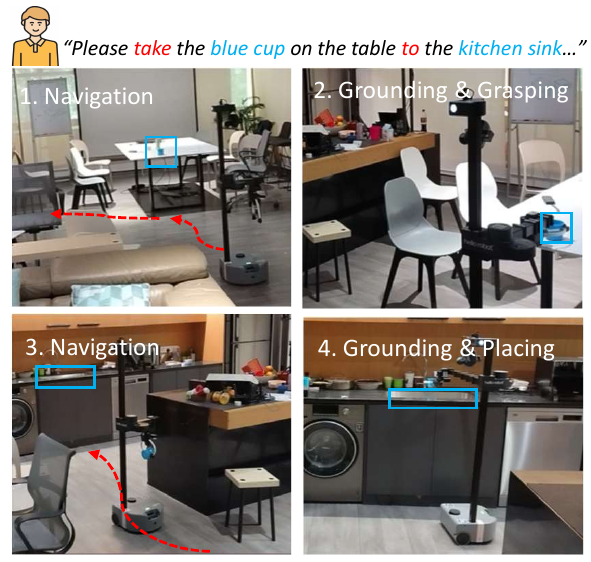}
\par\end{center}%
\end{minipage}
\vspace{-10pt}
\caption{A demonstration of real-world mobile manipulation task.}
\label{fig:demo}
\vspace{-10pt}
\end{figure}
\vspace{-10pt}

\begin{table}[h]
\caption{Evaluation of real-world mobile manipulation task.}
\vspace{-7pt}
\centering
\resizebox{\columnwidth}{!}{%
\begin{tabular}{l|c|c|c|c}
\hline
Methods & Nav. & Grounding \& Grasp & Place & Task \tabularnewline
\hline
OK-Robot~\cite{liu2024ok} & 11/32 & 4/16 & 3/16 & 0/10 \tabularnewline
DynaMem~\cite{liu2024dynamem} & 13/32 & 6/16 & 4/16 & 0/10 \tabularnewline
Dynam3D+OWLv2~\cite{wang2025dynam3d,minderer2023scaling} &
21/32 & 9/16 & 7/16 & 1/10 \tabularnewline
\OurModel{} (Ours) & \textbf{23/32} & \textbf{12/16} &
\textbf{11/16} & \textbf{3/10} \tabularnewline
\hline
\end{tabular}%
}
\label{tab:real_world}
\end{table}

To validate the real-world generalization capability of \gh{our} \OurModel{}, we conduct real-world mobile manipulation experiments \gh{in Table~\ref{tab:real_world}}. This task consists of sub-tasks including navigation, grounding and grasping the target, \gh{and} then placing it at a specified location. For example: “Please take the blue cup on the table to the kitchen sink, then bring the red cup next to the sink to this table.” For a fair comparison, all methods use 
Hello Robot Stretch 3 with the AnyGrasp model~\cite{fang2023anygrasp} for object grasping and placing. We design 10 task samples 
\gh{with} each comprising several sub-tasks. 
\gh{Our} \OurModel{} fully completed 3 tasks, \gh{and} significantly outperforming other baseline methods. This demonstrates its generalization ability on real-world long-horizon tasks. More visualizations are provided in the appendix.

\section{Conclusion}
\label{sec:conclusion}
This paper introduce D3D-VLP, a model 
\gh{that resolves} the trade-off between opaque end-to-end and disjunct modular systems in embodied AI. Our core innovation is the Dynamic 3D Chain-of-Thought (3D CoT), which serves as the core engine for task planning by unifying multi-step planning, 3D grounding, and navigation within a single autoregressive 3D-VLM. This is powered by a CoT Memory feedback loop 
\gh{that} is critical for stateful \gh{and} dynamic replanning. To overcome data scarcity, we propose the Synergistic Learning from Fragmented Supervision (SLFS) strategy 
\gh{for} effective training on a large-scale hybrid dataset. Our D3D-VLP demonstrates a significant step forward in the construction of high-performance embodied agents.
\\

\noindent\textbf{Limitations.} Our current SLFS leverages implicit gradient signals to guide the CoT pipeline on unlabeled data, while its autoregressive CoT generation exhibits limited exploratory behavior. Future work could incorporate Reinforcement Learning to further enhance this framework. By actively exploring the planning, grounding, and action spaces and optimizing against environmental rewards, RL may enable more adaptive reasoning policies and better exploit the potential of large-scale unlabeled data.
\clearpage
{
    \small
    \bibliographystyle{ieeenat_fullname}
    \bibliography{main}
}
\clearpage

\appendix

\section{Details of 3D CoT Dataset}
\label{sec:3d_cot_dataset} 

\begin{figure*}[hb]
\makebox[\textwidth][c]
{\includegraphics[width=0.7\paperwidth]{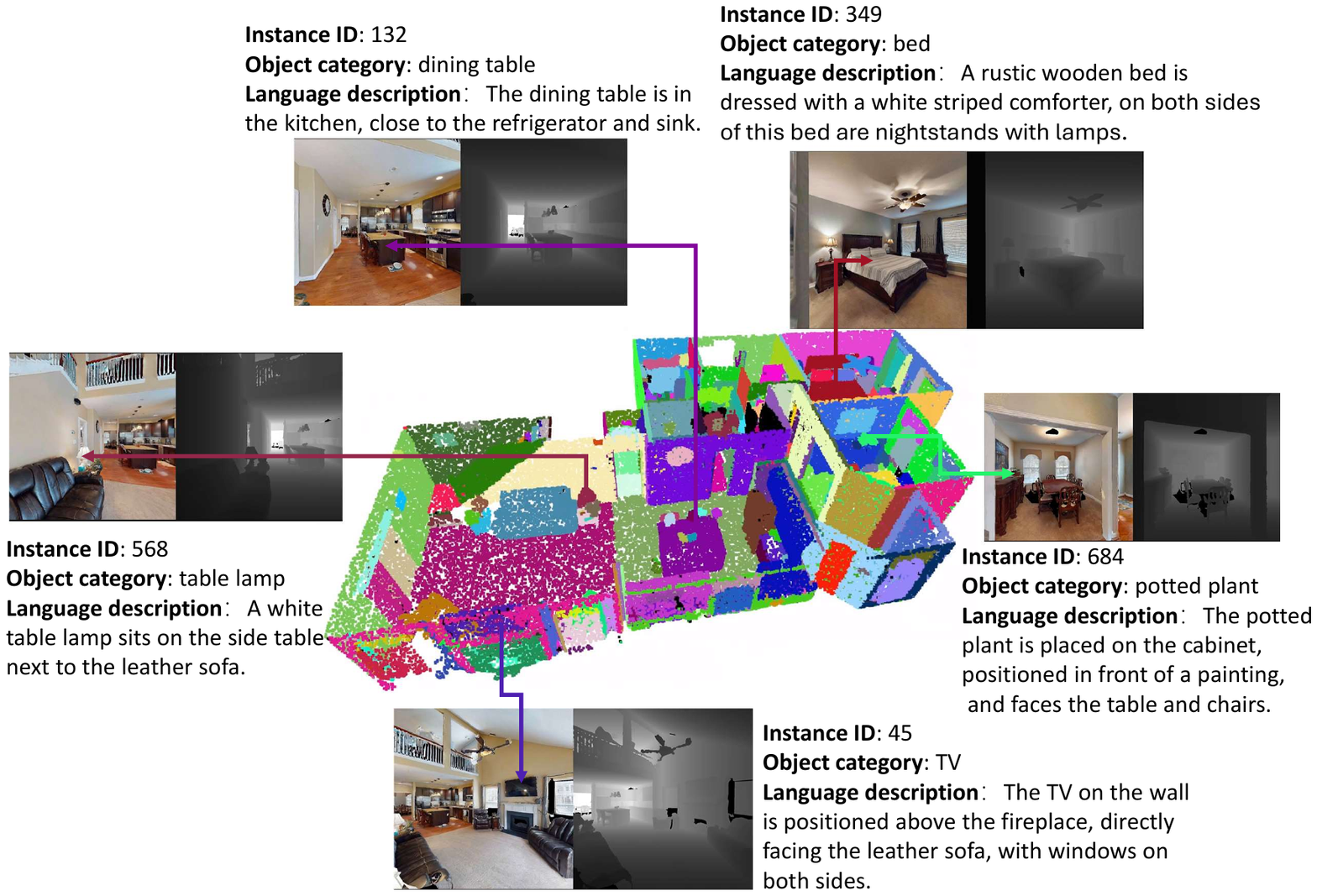}}
\vspace{-20pt}
\caption{Demonstration of a 3D scene in the training data~\cite{jia2024sceneverse,wang2024g3d}. Instance-level point clouds mark all instances with object categories and language descriptions.}
\label{fig:instance_pointcloud}
\vspace{-10pt}
\end{figure*}

\noindent\textbf{Grounding Annotations.}
We collect large-scale 3D instance-text pairs (Figure~\ref{fig:instance_pointcloud}) primarily from SceneVerse~\cite{jia2024sceneverse}, MMScan~\cite{mmscan}, PQ3D~\cite{pq3d}, and Grounded 3D-LLM~\cite{chen2024grounded}, as illustrated in Listing~\ref{lst:gn_example}.

\noindent\textbf{Navigation Annotations.}
We adapt instruction-following navigation data from R2R-CE~\cite{krantz2020beyond}, REVERIE-CE~\cite{qi2020reverie,wang2025dynam3d}, SRDF~\cite{wangbootstrapping}, and NavRAG~\cite{wang2025navrag} for the Habitat simulator environment~\cite{habitat,puig2023habitat3}.

\noindent\textbf{Grounding-Navigation Annotations.}
For open-vocabulary object grounding and navigation, we leverage instance-text annotations from HM3D~\cite{ramakrishnan2habitat,hm3d-sem} and MP3D~\cite{chang2017matterport3d} that support the Habitat simulator~\cite{habitat,puig2023habitat3}. We further integrate synthetic data from HSSD~\cite{khanna2024habitat}, ProcTHOR-10K~\cite{khanna2024habitat,procthor}, and ProcTHOR-Objaverse (AI2-THOR)~\cite{ehsani2024spoc,kolve2017ai2}. 
To generate trajectories for these instances:
\begin{itemize}
    \item \textbf{Habitat Simulator:} For HM3D, MP3D, HSSD, and ProcTHOR-10K (Habitat), the simulator computes optimal trajectories based on the scene mesh, ensuring collision-free paths with minimal distance.
    \item \textbf{AI2-THOR Simulator:} For ProcTHOR-Objaverse (AI2-THOR), as the native simulator~\cite{kolve2017ai2} lacks a sufficient ground-truth path computation function for our needs, we implemented an A* algorithm based on traversable BEV maps to generate shortest paths while avoiding obstacles.
\end{itemize}

\noindent\textbf{Planning-Grounding-Navigation Annotations.}
These comprehensive samples are primarily sourced from SG3D~\cite{sg3d} and Grounded 3D-LLM~\cite{chen2024grounded}. As shown in Listing~\ref{lst:pgn_example}, these annotations contain hierarchical instructions with multiple sub-goals. Each sub-goal specifies the target object category, its index in the instance point cloud (Figure~\ref{fig:instance_pointcloud}), and the center coordinates for the instance.

\noindent\textbf{Planning-Grounding Annotations.}
Scenes with planning annotations that are currently incompatible with the Habitat simulator, specifically ScanNet~\cite{dai2017scannet}, 3RScan~\cite{3rscan}, and ARKitScenes~\cite{baruch1arkitscenes}, are utilized for planning and grounding tasks. While these samples lack navigational trajectories, they provide rich instance-level grounding and planning instructions.

\noindent\textbf{Planning-Navigation Annotations.}
We incorporate data from VLN-Trans~\cite{zhang2023vln} (see Listing~\ref{lst:pn_example}). In these samples, the sub-goal coordinates correspond to trajectory waypoints rather than specific object instances, focusing on path planning and navigation without explicit object grounding.

\noindent\textbf{3D Scene Composition.}
The annotations described above span a wide range of environments, including posed RGB-D videos from real-world datasets (ScanNet~\cite{dai2017scannet}, 3RScan~\cite{3rscan}, ARKitScenes~\cite{baruch1arkitscenes}), high-quality real scans (HM3D~\cite{ramakrishnan2habitat}, MP3D~\cite{chang2017matterport3d}), and synthetic scenes (HSSD~\cite{khanna2024habitat}, ProcTHOR-10K~\cite{procthor}, ProcTHOR-Objaverse~\cite{spoc}).

\noindent\textbf{Grounding Label Assignment.} 
Unlike traditional 3D grounding tasks where all candidate objects are visible, the target object may be unobserved (i.e., lacking corresponding 3D tokens) during our online training. In such cases, the grounding output is assigned to a special \texttt{$<$grounding\_none$>$} token. When tokens corresponding to the target instance exist within the patch or instance tokens, we optimize the model using a multi-label cross-entropy loss, treating the target tokens as positive samples and the rest as negative. Following g3D-LF~\cite{wang2024g3d} and Dynam3D~\cite{wang2025dynam3d}, a 3D token is assigned to an instance based on its nearest neighbor in the ground-truth instance point cloud (available in HM3D, MP3D, ScanNet, 3RScan, and ARKitScenes). For scenes lacking instance point cloud annotations (e.g., HSSD, ProcTHOR-10K, and ProcTHOR-Objaverse), we only consider patch tokens within a 0.2m radius of the target instance center as positive samples.

\begin{figure*}[!ht]
    \centering
    \captionof{lstlisting}{\textbf{Planning-Grounding-Navigation Annotation.} An example showing the unified data structure and the corresponding prompt format.}
    \label{lst:pgn_example}
    \begin{lstlisting}[language=json]
// 1. JSON Annotation
{
  "scene_id": "hm3d/00378-DqJKU7YU7dA",
  "instruction": "Prepare for a shower.",
  "planning": [
    "1. Go to the shower containing a washcloth.",
    "2. Turn on the water to adjust the temperature.",
    "3. Take a towel from the rack stand to the left of the sink.",
    "4. Hang the towel on the shower curtain rod right by the shower."
  ],
  "instance_id": [[797], [797], [230], [515]],
  "instance_type": ["shower", "shower", "towel", "shower curtain rod"],
  // Coordinates truncated for display
  "instance_position": [
    [-6.554, -28.336, 0.863], 
    [-6.554, -28.336, 0.863],
    [-2.407, -14.418, 1.189], 
    [-3.649, -20.311, 0.981]
  ]
}

// 2. Input Prompt Trace
<|im_start|>system\n You are a helpful assistant<|im_end|\n
<|im_start|>user\n 
<patch tokens>\n <instance tokens>\n <zone tokens>\n 
The instruction: Prepare for a shower.\n
The history plans: 1. Go to the shower containing a washcloth.<shower_1>\n
The previous waypoints:<waypoint_4>,<waypoint_7>,<waypoint_9>,<waypoint_15>\n
The candidate waypoints: <waypoint_17><waypoint_18><waypoint_19>\n
Please give deep thinking plans. <|im_end|>\n<|im_start|>assistant\n
// 3. Output Trace
The next plans: 2. Turn on the water to adjust the temperature.\n
3. Take a towel from the rack stand to the left of the sink.\n
4. Hang the towel on the shower curtain rod right by the shower.\n
The grounded:target<shower_1>\n
The navigation action:waypoint<waypoint_19>reached the subgoal\n
<|im_end|>
\end{lstlisting}
\end{figure*}

\begin{figure*}[!ht]
    \centering
    \captionof{lstlisting}{\textbf{Planning-Navigation Annotation.} An example for vision-and-language navigation task with detailed step descriptions.}
    \label{lst:pn_example}
    \begin{lstlisting}[language=json]
{
  "scene_id": "mp3d/PX4nDJXEHrG",
  "instruction": "Walk across patio into the house. Walk forward toward wall with stone tile. Walk past stone tile wall on left side. Walk past stair case. Stop at entrance to kitchen area.",
  "planning": [
    "1. Walk across patio into the house.",
    "2. Walk forward toward wall with stone tile.",
    "3. Walk past stone tile wall on left side.",
    "4. Walk past stair case.",
    "5. Stop at entrance to kitchen area."
  ],
  "habitat_start_position": [-11.546, 0.115, 4.632],
  "habitat_start_rotation": [0, 0.998, 0, -0.049],
  "instance_id": [[-10000],[-10000],[-10000],[-10000],[-10000]],
  "instance_type": [null,null,null,null,null],
  "instance_position": [
    [-11.559, -3.040, 0.115],
    [-11.066, -1.661, 0.115],
    [-12.035, 1.826, 0.115],
    [-13.690, 4.130, 0.115],
    [-13.619, 5.133, 0.115]
  ]
}
    \end{lstlisting}

    \vspace{0.5em}

    \captionof{lstlisting}{\textbf{Grounding-Navigation Annotation.} An example for open-vocabulary object grounding and navigation task without explicit intermediate planning steps.}
    \label{lst:gn_example}
    \begin{lstlisting}[language=json]
{
  "scene_id": "hm3d/iigzG1rtanx",
  "instruction": "Please find the printer. The printer is next to the paper tray and is placed on desk.",
  "planning": "",
  "instance_id": [[96]],
  "instance_type": ["printer"],
  "instance_position": [
    [6.704, 1.147, 2.792]
  ]
}
    \end{lstlisting}
    \label{fig:annotation_examples_2}
\end{figure*}

\section{Details of Real-world Mobile Manipulation}
We validate the effectiveness of our D3D-VLP in real-world scenarios using the Hello Robot Stretch 3 mobile manipulator. The robot is equipped with a head-mounted Intel RealSense D435i RGB-D camera, which captures streaming posed RGB-D images. These streams are processed in real-time by the Dynam3D Encoder~\cite{wang2025dynam3d} to incrementally construct and update the Multi-level 3D Memory, ensuring the agent maintains a persistent and structured 3D scene representation during exploration.

For inference, we deploy the model on a remote workstation equipped with an NVIDIA RTX 4090 GPU and 64GB of RAM. The workstation handles the computationally intensive 3D-VLM reasoning and Dynamic 3D CoT generation, communicating with the robot via a local area network (LAN) over WiFi. Our deployment framework is adapted from DynaMem~\cite{liu2024dynamem}, which we extended to support our waypoint-based action space and local obstacle avoidance. The experiments are conducted in a physical home-like environment comprising a living room, kitchen, meeting room, and office. Crucially, to strictly evaluate zero-shot generalization, none of the objects or scene layouts in this environment are included in the training dataset.
\label{sec:mobile_manipulation}


\end{document}